\pdfoutput=1
\documentclass[sigconf]{acmart}
\setcopyright{acmcopyright}
\copyrightyear{2022}
\acmYear{2022}

\acmConference[MM'22]{30th ACM International Conference on Multimedia}{10 - 14 October
2022}{Lisbon, Portugal}
\acmBooktitle{MM '22: 30th ACM International Conference on Multimedia,
 10 - 14 October
2022, Lisbon, Portugal}
\acmPrice{15.00}
\acmISBN{978-1-4503-XXXX-X/18/06}

\acmSubmissionID{123-A56-BU3}

\usepackage{times}
\usepackage{epsfig}
\usepackage{graphicx}

\usepackage{booktabs,subcaption}
\usepackage{bbm}
\usepackage{adjustbox}
\usepackage{xcolor}
\usepackage{multirow}

\usepackage{amsmath,amsfonts,bm}

\def\eqref#1{equation~\ref{#1}}

\def\1{\bm{1}}

\def\vz{{\bm{z}}}

\def\mA{{\bm{A}}}

\def\mF{{\bm{F}}}

\DeclareMathAlphabet{\mathsfit}{\encodingdefault}{\sfdefault}{m}{sl}
\SetMathAlphabet{\mathsfit}{bold}{\encodingdefault}{\sfdefault}{bx}{n}

\def\gG{{\mathcal{G}}}

\def\sR{{\mathbb{R}}}

\newcommand{\R}{\mathbb{R}}

\renewcommand{\eqref}[1]{(\ref{#1})}
\newcommand{\eqqref}[1]{Eq.~(\ref{#1})}
\newcommand{\figgref}[1]{Fig.~\ref{#1}}

\newcommand{\ours}{\textsc{\textbf{CVGAE} }}
\newcommand{\auc}{AUC }

\newcommand {\eg}{\emph{e.g.}} 
\newcommand {\ie}{\emph{i.e.}} 
\newcommand\etal{\emph{et al.}}

\usepackage{float}
\usepackage{pifont}
\usepackage{dsfont}

\newcommand{\myparagraph}[1]{\vspace{2pt}\noindent{\bf{#1}}}

\usepackage{siunitx}
\begin{document}

\title{On Leveraging Variational Graph Embeddings for  \\ Open World Compositional Zero-Shot Learning}

\author{
Muhammad Umer Anwaar $^{1,2}$, Zhihui Pan$^{1,2}$, Martin Kleinsteuber$^{1,2}$\\
${^1}$Technische Universit{\"a}t M{\"u}nchen, Germany\\
${^2}$Mercateo AG, Germany\\
\{umer.anwaar, zhihui.pan\}@tum.de, martin.kleinsteuber@mercateo.com
}

\renewcommand{\shortauthors}{Anwaar et al.}

\begin{abstract}
Humans are able to identify and categorize novel compositions of known concepts. The task in Compositional Zero-Shot learning (CZSL)
is to learn composition of primitive concepts, i.e. objects and states, in such a way that even their novel compositions can be zero-shot classified.
In this work, we do not assume any prior knowledge on the feasibility of novel compositions \ie open-world setting, where infeasible compositions dominate the search space.
We propose a \textbf{C}ompositional \textbf{V}ariational \textbf{G}raph \textbf{A}utoencoder (\textbf{CVGAE}) approach for learning the variational embeddings of the primitive concepts (nodes) as well as feasibility of their compositions (via edges).
Such modelling makes CVGAE scalable to real-world application scenarios.
This is in contrast to SOTA method, CGE \cite{cge}, which is computationally very expensive.
\eg for benchmark C-GQA dataset, CGE requires \num{3.94e5} nodes, whereas CVGAE requires only 1323 nodes.
We learn a mapping of the graph and image embeddings onto a common embedding space.
CVGAE adopts a deep metric learning approach and learns a similarity metric in this space via
bi-directional contrastive loss between projected graph and image embeddings.
We validate the effectiveness of our approach on three benchmark datasets.
We also demonstrate via an image retrieval task that the representations learnt by CVGAE are better suited for compositional generalization.
\end{abstract}


\begin{teaserfigure}
  \includegraphics[width=\textwidth]{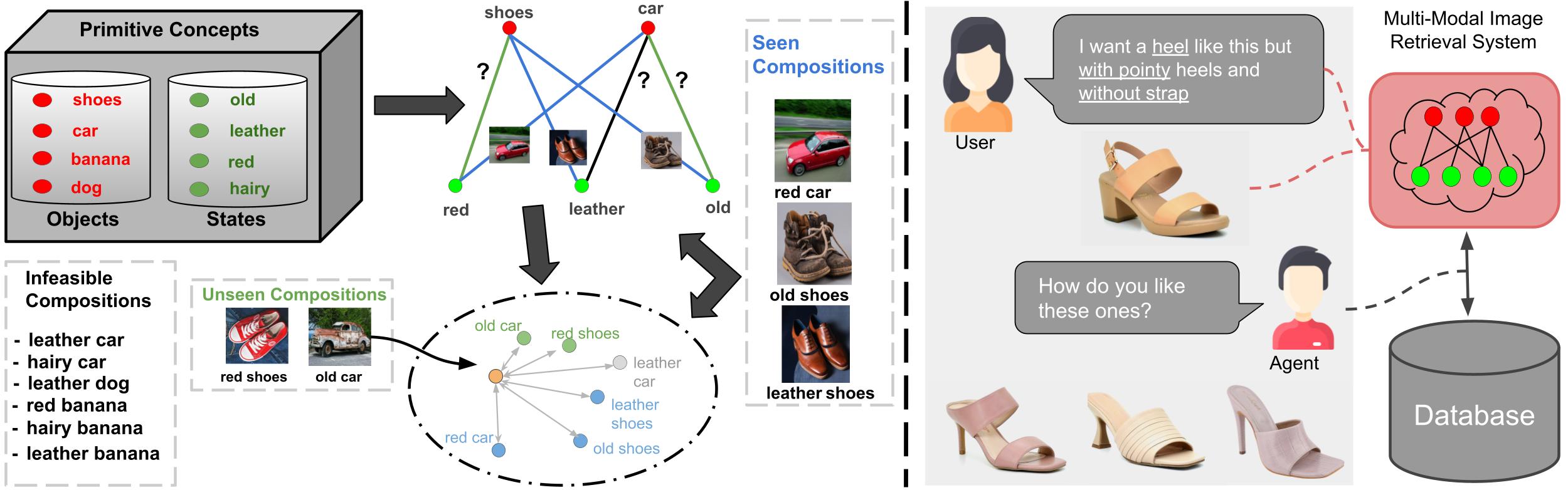}
  \caption{\textbf{Left}: Open World CZSL Task. Most compositions of primitive concepts in the model search space are infeasible. \textbf{Right}: A potential application scenario, where CZSL aids the image retrieval system in addressing the multi-modal query.}
  \label{fig:teaser}
\end{teaserfigure}

\maketitle
\thispagestyle{empty}

\section{Introduction}
In their seminal works in vision and cognitive science, 
Biederman \cite{PsychologicalBiederman} and Hoffman \etal \cite{CognitionHoffman1984PartsOR} showed the compositional nature of human perception of the visual world.
This compositional nature is the motivation behind the feature compositionality which 
has been exploited by modern vision systems.
For instance, 
learning image features \cite{zeiler2014visualizing, lecun1989backpropagation} and transfer learning \cite{choi2013adding, deng2014large, patricia2014learning}.
It is well-known that unfamiliar compositions of concepts dominate the long tailed distribution of visual concepts \cite{salakhutdinov2011learning, wang2017learning}.
However, the bulk of the research is focused on composition in feature space, whereas composition of concepts in the classifier space has received less attention.

Recently, several works have focused on the so-called Compositional Zero-Shot Learning (CZSL) problem \cite{misra2017redwine,purushwalkam2019tmn,li2020symnet}. The task is to learn the composition of concepts (objects and states) in such a way that even their novel compositions can be classified without any supervision. 
Interestingly, 
the baseline and state-of-the-art (SOTA) methods in CZSL assume that the
novel composition of concepts (objects and states) in the model output space is already known at test time.
The only exception is the work by Mancini \etal \cite{compcos}, which do not impose any such constraint on the model output space. Following \cite{compcos}, we refer to the former as Close-World (CW) CZSL task and the latter as Open-World (OW) CZSL task.
The left part of Fig.~1 presents the OW-CZSL task, whereas the right part presents a potential application scenario where OW-CZSL can help E-commerce websites.
It is to be noted that underlined object and states can be recognized via Part-of-speech (POS) tagging \cite{pos1,pos2}, but this is out of scope of this work.

In this work, we propose \ours, a Variational Graph Autoencoder (VGAE) based approach for tackling both CW and OW CZSL tasks in a principled way. 
We argue that objects and states are being generated from a prior distribution. They are treated as nodes of a graph and an edge between them indicates the existence of a compositional pair.
This formulation enables us to learn the latent representation of our primitive concepts in a space of reduced dimensionality along with the feasibility of their compositions (edges). 
The embeddings of the compositional pairs are obtained by simply concatenating the respective state and object node embeddings.
This is in contrast to CGE\cite{cge}, which requires nodes as well as all compositional pairs in the graph. 
Such problem formulation by Naeem \etal\cite{cge} limits its applicability for the real-world scenarios.
We discuss the computational complexity in detail in \autoref{sec: time}.

We hypothesize that there are two fundamental obstacles in learning models which exhibit compositional generalization, \ie (1) zero-shot nature of unseen labels (distribution-shift) and (2) learning the ``disentanglement" of primitive concepts from training samples.
In this paper, we do not claim to overcome these challenges. Rather, we recognize these obstacles and draw inspiration from them in developing our model.
This is the reason for choosing a generative process for the primitive concepts in our graph.
We hypothesize that keeping only the primitives in the graph aids in learning of primitive concepts in a disentangled way. Moreover, the variational modelling process also ensures that \ours learns the underlying distribution rather than learning such spurious correlations which hurt inference at test time.
This enables \ours to achieve better compositional generalization.

Our model focuses on learning the distribution of the primitive concepts and predicting the existence (feasibility) of an edge between them by exploiting the graph information.
We argue that once good primitive representations are learned in a disentangled way,
it is more likely to be stable across training and test compositional labels. 
Following the literature \cite{cge,compcos,li2020symnet}, we employ a pretrained image model to extract image features.
We learn a mapping of the compositional pair embeddings and image features on the common embedding space.
In this space, we employ contrastive loss to learn embeddings which are visually grounded and semantically meaningful.
That is, by maximizing the similarity between image embeddings with target compositional pair embeddings against random pairs and vice-versa through a bidirectional contrastive loss.
Another big advantage of \ours is that we exploit the graph information 
and augment the supervised contrastive loss with VGAE loss. 
This is in contrast to the current SOTA methods, which overwhelmingly rely on supervisory labels for learning. 
Such purely supervised methods disregard the fact that the underlying data lives on a much complicated manifold than what sparse labels could capture.
This often leads to good task-specific solutions, rather than learning the multiple semantic concepts in the data.
We argue that representations learnt by \ours better capture the semantic meaning of primitive concepts.
We evaluate the validity of our approach via experiments on three benchmark datasets (See Sec.~\ref{sec:experiment}).
Our main contributions are summarized below: 
\begin{itemize}

\item We propose \ours, a variational graph-based approach for tackling both CW and OW CZSL tasks. 
We argue that the node embeddings and edges between the primitive concepts (object and states) are sufficient for 
achieving good compositional generalization.

\item We show that \ours is computationally cheaper than the current graph-based SOTA method, CGE.


\item \ours achieves better compositional generalization than SOTA methods on three benchmark datasets, \ie for OW-CZSL task, the performance gain with respect to best harmonic mean metric is 12.5\% on MIT-States, 8.3\% on UT-Zappos and 25\% on C-GQA dataset.
\end{itemize}

\section{Related Work}

\subsection{Compositional Zero-Shot Learning}

Early works in CZSL see state (attribute) as a mid-level feature between visual patterns and object recognition. 
It has been proved to help various tasks, including zero-shot recognition, image description~\cite{liu15,kumar08,farhadi09}, visual question answering~\cite{koushik2017} and so on. \cite{farhadi09,jayaraman14} treat object as combination of states (attributes). \cite{jayaraman14ua} develop a random forest algorithm on top of attribute classifiers. \cite{parikh11} propose using continuous instead of binary state classifier and infer to unseen objects by relating its description to seen objects. Skip-gram word embeddings can be used for both state and object to make predictions on unseen relations~\cite{alhalah16}.

Instead of being used as an intermediate feature, attribute classification becomes another goal of learning in compositional learning, besides object classification. ~\cite{chen2014inferring} uses matrix factorization to get unseen object-specific attribute classifiers from seen ones. ~\cite{misra2017redwine} combine attribute and object classifier via a transformation network to get the composition classifier. Instead of training classifiers for primitives, \cite{purushwalkam2019tmn} use the state and object label to control a gated network calculating a triplet loss of (image, state, object). 

Besides the visual presentation, the semantic information of primitives are also utilized to generalize to unseen compositions. \cite{nagarajan2018attributeasoperators} models object using its semantic representation and attribute as linear transformation for the object. \cite{li2020symnet} learns the coupling and decoupling mechanism between attributes and objects.

\subsection{Open World Recognition}

Most previous works conduct their experiments on the closed-world setting, where they assume prior knowledge of all test  compositions.
This weakens their generalizability to practical settings. We adopt the more realistic open-world setting from ~\cite{compcos}. The search space now contains every possible combination of state and object. 
We show that most previous models exhibit performance degradation in this setting, while our model can handle this new problem well with the help of Graph Neural Network(GNN). 

Both our work and CGE~\cite{cge} use GNN to get the desired composition representation. 
However, we propose a variational graph approach which requires much less nodes. This is especially advantageous in the open-world setting considering the already enormous search space. 

\section{Methodology}
\subsection{Task Formulation} \label{task}

Let $\mathcal{X}$ denote the set of images, $\mathcal{S}$ denote the set of states, $\mathcal{O}$ the set of objects and $\mathcal{Y} = \mathcal{S}\times\mathcal{O}$ denote the set of all the compositions.
The set $\mathcal{Y}$ can also be expressed as a union of real compositional labels, denoted as $\mathcal{Y}_r$, and hypothetical compositional labels, denoted as $\mathcal{Y}_h$. \ie $\mathcal{Y}=\mathcal{Y}_r\cup\mathcal{Y}_h$.
$\mathcal{Y}_r$ consists of the ``real" labels, \ie we have images in the dataset for these labels.
$\mathcal{Y}_r$ is partitioned into two disjoint subsets, depending on whether the corresponding label was ``seen" by the model during training or not. We refer to them as ``seen" $\mathcal{Y}_s$ and ``unseen" $\mathcal{Y}_u$ respectively. 
\ie $\mathcal{Y}_r=\mathcal{Y}_s\cup\mathcal{Y}_n$ and $\mathcal{Y}_s\cap \mathcal{Y}_n=\emptyset$.
We can write the training set
as: $\mathcal{T}=\{(x, y) | x\in \mathcal{X}, y\in\mathcal{Y}_s \}$, where $\mathcal{Y}_s\subset\mathcal{Y}_r$.

In this work, following \cite{cge, compcos, purushwalkam2019tmn, zslxian18benchmark}, we adopt the generalized compositional zero-shot split where the test set includes images from both $\mathcal{Y}_s$ and  $\mathcal{Y}_u$.
The goal of the model $f:\mathcal{X}\rightarrow \mathcal{Y}_{\xi}$ is to predict compositional labels in a space $\mathcal{Y}_{\xi}\subseteq\mathcal{Y}$.
Depending on the output space $\mathcal{Y}_{\xi}$ of the model, we get two variants of this task: (1) Close-World (CW) CZSL \ie $\mathcal{Y}_{\xi}\equiv\mathcal{Y}_r$ and (2) Open-world (OW) CZSL \ie $\mathcal{Y}_{\xi}\equiv\mathcal{Y}$.
In other words, in CW CZSL, the set of unseen compositions $\mathcal{Y}_u$ is assumed to be known a priori. 
Thus, we consider $\mathcal{Y}_h=\emptyset$. 
Whereas, OW CZSL does not have this assumption, which makes it a more challenging task and closer to real-world application scenarios.

\subsection{Proposed Method: CVGAE} \label{model}

Suppose {an undirected} and unweighted graph $\gG$ consisting of $N$ nodes, with adjacency matrix $\mA \in \sR^{N\times N}$ and a matrix $\mF \in \sR^{N \times m}$ of $m$-dimensional node features.
For \ours, $N=|\mathcal{S}| + |\mathcal{O}|$ and each row in $\mF$ consists of word embedding of the corresponding state or object. 
This is in contrast to CGE \cite{cge}, which requires an additional $|\mathcal{Y}_\xi|$ nodes, corresponding embeddings and edges. 
Such formulation makes CGE computationally very expensive than our approach (see \autoref{sec: time}).

\ours jointly encodes the information in nodes and edges in a $h$ dimensional latent space. Let us denote the latent node embedding by $\vz_{i}  \in \R^h$.  
We assume that these latent random variables $\{\vz_{1}, \bm{z_2}, \cdots, \bm{z_N}\}$ follow standard gaussian distribution. 
These latent variables are stacked into a matrix $\bm{Z} \in \mathbb{R}^{N \times h}$.
The joint distribution can be mathematically expressed as
\setcounter{figure}{1}

{\small
\begin{equation}
	p(\gG, \bm{Z})  = p( \bm{Z}) p_{\theta}(\mathcal{G}| \bm{Z}),
\end{equation}
}%
{\small
\begin{align}
	p(\bm{Z})   & = \prod \limits_{i=0}^{N} p(\bm{z}_i), \label{eq:vgae-pz} \\                    
	p(\bm{z}_i) & = \mathcal{N}(\bm{0}, \mathrm{diag}(\bm{1})) \ \forall i .
	\label{eq:vgae-pz_i} 
\end{align}
}%
\vspace{-1.25em}

\begin{figure*}
\centering
\includegraphics[width=0.99\linewidth]{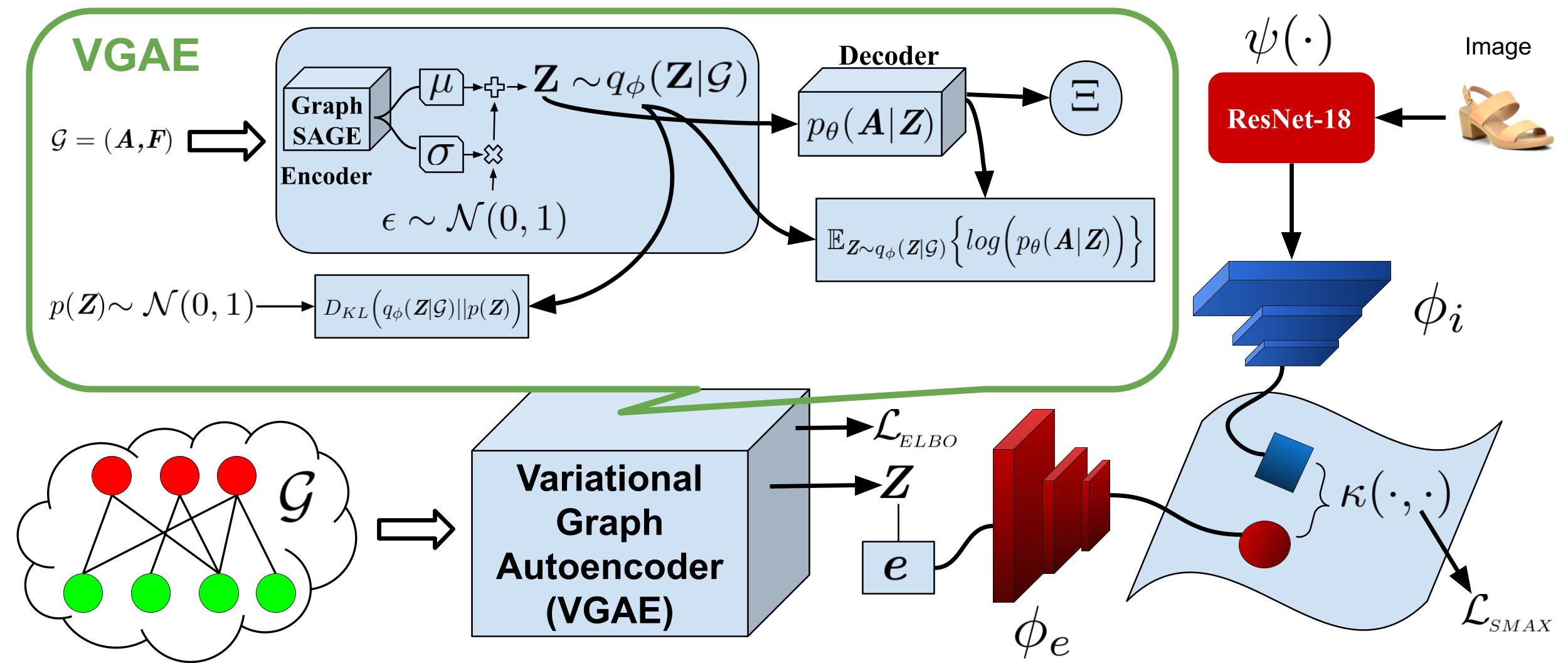}
\caption{\ours: a variational graph-based approach for tackling both CW and OW CZSL tasks. The key idea is that the node embeddings and edges between the primitive concepts (object and states) are sufficient for 
achieving good compositional generalization.
}
\label{CVGAE_Architecture}
\end{figure*}

For an unweighted and undirected graph $\mathcal{G}$, we follow \cite{vgae} and restrict the decoder to reconstruct only edge information from the latent space.
The edge information can be represented by an adjacency matrix $\bm{A} \in \mathbb{R}^{N \times N}$ where {$a_{ij}$} refers to the element in $i^{th}$ row and $j^{th}$ column. If an edge exists between node $i$ and $j$, we have  {$a_{ij} = 1$} else $0$. 
Thus, the decoder is given by
{\small
\begin{align}
	p_{\theta}(\bm{A}| \bm{Z}) = \prod \limits_{(i,j)=(1,1)}^{(N,N)}p_{\theta}(a_{ij} = 1| \bm{z_i}, \bm{z_j}), \label{eq:pA-given-Z}
\end{align}
}%
\vspace{-1.25em}
{\small
\begin{align}
	p_{\theta}(a_{ij} = 1| \bm{z_i}, \bm{z_j}) = \sigma(<\bm{z_i}, \bm{z_j}>), \label{edgedecoder}
\end{align} 
}%

where $<\cdot,\cdot>$ denotes dot product and $\sigma(\cdot)$ is the logistic sigmoid function.
It is to be noted that an additional restriction is imposed when sampling the nodes, \ie $\bm{z_i} \in \mathcal{S} $ and $\bm{z_j} \in \mathcal{O}$.
In order to ensure computational tractability, we introduce the approximate posterior. \ie
{\small
\begin{align}
	q_{\phi}(\bm{Z} | \mathcal{G})         & = \prod \limits_i^N q_{\phi}(\bm{z_i} | \mathcal{G}) \label{eq:vgae-qz_given_G} \\
	\quad q_{\phi}(\bm{z_i} | \mathcal{G}) & = \mathcal{N}\Big(\bm{\mu_i}(\mathcal{G)}, \mathrm{diag}(\bm{\sigma_i^2}(\mathcal{G}))\Big) 
	\label{eq:vgae-qz_i_given_G}
\end{align}
}%

{where $\gG = (\mA, \mF)$.
We employ GraphSAGE \cite{GraphSAGE}} as node encoder to learn
$\bm{\mu}_i(.)$ and   $\bm{\sigma}_i(.)$
Reparameterization trick \cite{vae} is used to obtain samples of $q_{\phi}(\bm{Z} | \mathcal{G})$ from the mean and variance.

The training objective should be such that the model is able to generate new data and recover graph information 
from the embeddings simultaneously.
We maximize the corresponding ELBO bound (for derivation, refer to the supplementary material), given by
{\small
\begin{align}
	\mathcal{L}_{\mathrm{ELBO}}  & = \sum \limits_{i=1}^{N}D_{KL}\bigg(\mathcal{N}\Big(\bm{\mu}_i(\mathcal{G}), \bm{\sigma}^2_i(\mathcal{G})\Big) \ || \  \mathcal{N}(\bm{0}, \mathrm{diag}(\bm{1})) \bigg) \nonumber \\
	  &  
	  - \mathbb{E}_{\bm{Z} \sim q_{\phi}(\bm{Z} | \mathcal{G})} \Big\{ log\Big( p_{\theta}(\bm{A}| \bm{Z})\Big)\Big\}
	  \label{eq:vgae-final-loss} 
\end{align}
}%

where $D_{KL}$ denotes the Kullback-Leibler (KL) divergence and the second term is the binary cross-entropy loss between input edges and the reconstructed edges. 
\subsection{Network Architecture}
Fig. \ref{CVGAE_Architecture} presents the overview of the architecture of \ours, a variational graph based approach for learning composition of multiple semantic concepts in images.
Here, $\psi(\cdot)$ denotes the pre-trained image model (e.g. ResNet-18), which takes an image as input and returns image features $\bm{x}$ in a $d$-dimensional space.
We get the embeddings for all the compositional pairs of interest $\mathcal{Y}_\xi$ by concatenating respective state and object node embeddings. \ie for each $y \in \mathcal{Y}_\xi$, we have the pair embeddings $\bm{e} = concat (\bm{z},\bm{z'})$, where $\bm{z} \in \mathcal{S} $ and $\bm{z'} \in \mathcal{O}$.
We project these embeddings onto a common multi-modal embedding space via separate projection modules. 
These modules are denoted by $\phi_e: \mathbb{R}^{2h} \mapsto \mathbb{R}^k$ and $\phi_i: \mathbb{R}^d \mapsto \mathbb{R}^k$ for compositional pair and image respectively. 

In this common embedding space, we aim to learn such representations which better capture the underlying cross-modal dependencies. 
Our aim is to maximize the similarity between the projected compositional pair embedding $\phi_e(\bm{e})$ and the corresponding projected image features $\phi_i(\bm{x})$.
We adopt a deep metric learning (DML) approach to train \ours. 
Let $\kappa(\cdot,\cdot)$ denote the similarity kernel, which we implement as a dot product between its inputs.
The objective is to learn a similarity metric, $\kappa(\cdot,\cdot): \mathbb{R}^k \times \mathbb{R}^k \mapsto \mathbb{R}$, between $\phi_e(\bm{e})$ and $\phi_i(\bm{x})$.

In other words, \ours should learn to map semantically similar points from the data manifold in $\mathbb{R}^{2h} \times \mathbb{R}^d$ onto metrically close points in $\mathbb{R}^k$.
Analogously, the model should push the projected image features away from the wrong compositional pair embeddings in $\mathbb{R}^k$.
For each training sample $(x^{(j)},y^{(j)})$ in the mini-batch of size $B$, we normalize the similarity between the projected compositional pair embedding $\phi_e(\bm{e}^{(j)})$ and the corresponding projected image features $\phi_i(\bm{x}^{(j)})$
by dividing it with the sum of similarities between $\phi_i(\bm{x}^{(j)})$ and all the projected compositional pair embeddings.
We denote this loss as $\mathcal{L}_{e\mapsto i}$.
\begin{align}
\label{eq:loss_ei}
\mathcal{L}_{e\mapsto i} =  \frac{1}{B}  \sum_{j=1}^B - \log \Bigg \{\frac{\exp\{\kappa(\phi_e(\bm{e}^{(j)}), \phi_i(\bm{x}^{(j)})\}}
{\sum_{r=1}^{|\mathcal{Y}_\xi|} \exp\{\kappa(\phi_e(\bm{e}^{(r)}), \phi_i(\bm{x}^{(j)})\}} \Bigg\},
\end{align}
In the other direction, we also divide this similarity with the sum of similarities between $\phi_e(\bm{e}^{(j)})$ and all the images in the batch. We denote this loss as $\mathcal{L}_{i\mapsto e}$.
\vspace{-1em}

 \begin{table*}[!h]
\centering

\scalebox{0.9}
{\begin{tabular}{l|c|c|c}
\toprule
& MIT-States & UT-Zappos & C-GQA \\
\toprule
\# pairs in output space \ie $|\mathcal{Y}_{\xi}|$ for CW/OW & 1962 / 28175 & 116 / 192 & 9378 / 394110 \\
Percentage of hypothetical pairs in OW $|\mathcal{Y}_h|/|\mathcal{Y}_\xi|$
& 92.7\% & 39.6\% & 97.6\%
\\
\# states/objects  & 115 / 245 & 16 / 12 & 453 / 870 \\
\# images (train/val/test) & 30k/10k/13k & 23k/3k/3k & 26k/7k/5k \\
\# seen pairs (train/val/test) & 1262/300/400 & 83/15/18 & 6963/1173/1022 \\
\# unseen pairs (val/test) & 300/400 & 15/18 & 1368/1047 \\
\bottomrule
\end{tabular}}
\vspace{3mm}
\caption{Dataset statistics for CZSL Task}
\label{tab:Datasets}
\end{table*}


\begin{align}
\label{eq:loss_ie}
\mathcal{L}_{i\mapsto e} =  \frac{1}{B}  \sum_{j=1}^B - \log \Bigg \{\frac{\exp\{\kappa(\phi_e(\bm{e}^{(j)}), \phi_i(\bm{x}^{(j)})\}}
{\sum_{b=1}^B \exp\{\kappa(\phi_e(\bm{e}^{(j)}), \phi_i(\bm{x}^{(b)})\}} \Bigg\},
\end{align}

The total loss can be written as:
\begin{align}
\label{total_loss}
\mathcal{L} = \mathcal{L}_{ELBO} +   \lambda_{ei} \  \mathcal{L}_{e\mapsto i} +   \lambda_{ie} \  \mathcal{L}_{i\mapsto e}.
\end{align}



\subsection{Feasibility Score}
Mancini \etal \cite{compcos} propose two variants of their model CompCos, with separate loss functions for the CW-CZSL and OW-CZSL scenarios. 
The main difference is that in the OW-CZSL setting, they employ a \textit{``feasibility score"} of the unseen compositional pairs in the loss function.
In contrast to their approach, \ours gets the \textit{feasibility score}, denoted by $\Xi$, of the unseen compositional pairs as a \textit{byproduct} of using the 
VGAE. 
This is because our approach makes use of both the structural information existing in the graph and the variational modelling of the nodes.
\ie, \ours employs an edge decoder, see \eqqref{edgedecoder}, to reconstruct the  edges of the graph. 
This enables us to leverage the graph structure of states and objects in such a way that ``edges" can be predicted between them via a simple dot product of their embeddings. 
If an edge (link) is predicted between a state-object pair, then it is treated as a proxy of the feasibility of their composition.
This major advantage arises naturally due to VGAE and we do not need to change our loss function to incorporate \textit{feasibility scores}. 
They are simply calculated during inference time and we employ threshold ($\tau$) to make a hard decision on the feasibility of a compositional pair. 
\begin{equation}
    \Xi = \mathbbm{1}(\bm{z}\bm{z'}^T \geq \tau) , \quad \tau \in [0,1].
\end{equation}
The threshold $\tau$ is calculated using validation set.
But for all our experiments we fixed $\tau=0.2$. Since the performance is quite similar for a long range of $\tau$, \ie from 0.05 to 0.3.


\section{Experiments} \label{sec:experiment}

\subsection{Experimental Setup}
In our experiments, we use three  benchmark datasets,  namely: MIT-States\cite{isola2015mitstates}, UT Zappos \cite{yu2017zappos} and C-GQA \cite{cge}. 
MIT-States consists of 53753 diverse real-world images where each image is described by an object-attribute composition label, i.e. an attribute (state) and a noun (object), e.g. “broken glass”. 
UT-Zappos is a dataset of only shoes with fine-grained annotations. 
A composition label consists of shoe type-material pair. 
Recently, Naeem \etal \cite{cge} proposed C-GQA dataset which is built on Stanford GQA dataset. This dataset has the largest label space among the publicly available CZSL datasets. 
Table \ref{tab:Datasets} presents the detailed statistics of the three datasets.
Following \cite{chao2017empirical, purushwalkam2019tmn, cge, compcos, li2020symnet}, we use the
Generalized CZSL (GCZSL) split of these benchmark datasets.
In this split, the model performance
is evaluated on both seen and unseen pairs.
For the image retrieval task, we follow the same train-test split and evaluation protocol as used by ComposeAE \cite{anwaar2020compositional}, the SOTA method for this task.
\ie the split ensures that there is no overlap between training and testing queries in terms of objects.

\myparagraph{Baselines:}
We compare the results of \ours with several baselines as well as state-of-the-art (SOTA) methods.

\noindent\textbf{AoP}~\cite{nagarajan2018attributeasoperators} employs GloVe vectors \cite{pennington2014glove} for objects. 
They model states (attributes) as linear transformation matrices (``operators") and their product with the object embeddings yields corresponding pair embeddings. 
The compositional pair with the minimum distance to image embedding is considered the prediction of the model.

\noindent\textbf{LabelEmbed (LE+)}  
was introduced by \cite{nagarajan2018attributeasoperators}. It employs Glove vectors for both state and object. It projects the image, state and object vectors into a common semantic space. The joint embeddings are then fed to the classifier.

\noindent\textbf{Task-driven Modular Networks (TMN)}~\cite{purushwalkam2019tmn} modifies a set of modules (fully connected layers operating in semantic concept space) through a gating function.
They propose to re-weight these primitive modules for generalizing to unseen compositions.

\noindent\textbf{SymNet}~\cite{li2020symnet} learns object embeddings showing symmetry under different state transformations. They propose that such symmetry constraints enable learning better embeddings for the CZSL task.

\noindent\textbf{CompCos}~\cite{compcos} proposes that the cosine similarity among objects and states can be used as a proxy to estimate the feasibility of each unseen composition.
They argue that similar objects share similar states whereas dissimilar objects are less likely to share similar states.

\noindent\textbf{CGE}~\cite{cge} learns a compatibility function between image features and classes of seen and unseen compositions from a graph. The graph consists of not only the primitive concepts but also the compositional pairs of interest. 

\myparagraph{Implementation Details:}
It has been recently shown that improvements in reported results in deep metric learning are exaggerated and performance comparisons are done unfairly \cite{musgrave2020metric}.
In our experiments, we took special care to ensure fair comparison. 
We follow the same evaluation protocol and metrics as current SOTA methods \cite{cge,compcos,purushwalkam2019tmn}.
Following \cite{compcos,li2020symnet,nagarajan2018attributeasoperators}, all the methods considered in this paper use 512-dimensional image features extracted by ResNet18 model pretrained on ImageNet~\cite{deng2009imagenet}. 
Resnet18 model is not fine-tuned on the CZSL datasets due to fairness concerns for the baseline methods \ie since the baseline methods report results with fixed ResNet18 model, thus it would be unfair to report the results of baselines without searching and choosing the optimum learning rate of the image model for the baselines. 
We employ the same initial graph embeddings as CGE \cite{cge}, \ie word2vec \cite{word2vec} for UT-Zappos and C-GQA; and concatenation of word2vec and fasttext\cite{fasttext} embeddings for MIT-States.
We use the validation set to determine the hyper-parameters,
\eg, learning rate, weights of the losses etc.
We employ Adam~\cite{kingma2014adam} optimizer with learning rate of $5e^{-5}$.
The weights of the losses are: $\lambda_{ei}=10$, 
and
$\lambda_{ie}=0.01$. 
We repeat each experiment 10 times and report the average performance of the models. 


\myparagraph{Metrics}
We follow the same evaluation protocol and metrics as current SOTA methods \cite{cge,compcos,purushwalkam2019tmn}.
For the task of GCZSL, we follow the evaluation protocol from TMN \cite{purushwalkam2019tmn} and use the same metrics as proposed by them. Namely: best accuracy on only images of seen/unseen compositions (\textit{best seen/best unseen}), best harmonic mean (\textit{best HM}) and Area Under the Curve (AUC) for seen and unseen accuracies by varying the bias values.
For the task of image retrieval, following the evaluation protocol from TIRG~\cite{TIRG}, we use recall at rank $k$ ($R@k$), as our evaluation metric. 
$R@k$  estimates the proportion of queries where the target (ground truth) image is within the top $k$ retrieved images.

{
\setlength{\tabcolsep}{2pt}
\renewcommand{\arraystretch}{1.5}
\begin{table}[!t]
\renewcommand{\tablename}{}
\renewcommand\thetable{(a)} 
        \begin{subtable}[h]{1\linewidth}
        \centering
        \resizebox{1\linewidth}{!}
        {
        \begin{tabular}{l|cc|cccc|cccc}
        \toprule
        \multicolumn{1}{l}{\textbf{Method}} & \multicolumn{2}{c}{\textbf{Concepts}} & \multicolumn{4}{c}{\textbf{Close-World}} &
        \multicolumn{4}{c}{\textbf{Open-World}}\\
        \toprule
        & State & Object  
        &Seen & Unseen  & HM & AUC 
        &Seen & Unseen  & HM & AUC \\
        \toprule
        AoP &  21.1 & 23.6 & 
        14.3 & 17.4  & 9.9 & 1.6 &
        16.6 & 5.7 & 4.7 & 0.7 \\
        LE+ & 23.5 & 26.3 &
        15.0 & 20.1 & 10.7 & 2.0 &
        14.2 & 2.5 & 2.7 & 0.3 \\
        TMN & 23.3 & 26.5 &
        20.2 & 20.1 & 13.0 & 2.9 &
        12.6 &  0.9 &   1.2 &   0.1 \\
        SymNet & 26.3 & 28.3 &
        24.4 & 25.2 & 16.1 & 3.0 &
        21.4   &7.0 &5.8   &0.8 \\
        CompCos & 27.9 & 31.8 & 
        25.3 & 24.6 & 16.4 & 4.5 &
        25.4 &\textbf{10.0} &8.9   &1.6 \\
        CGE & \textbf{27.9} & 32.0 &
        \textbf{28.7} & 25.3 &  17.2 & 5.1 &
        25.6 & 9.3 & 8.7 & 1.5 \\
        \bottomrule
        \textbf{\ours} &  25.6 & \textbf{32.3} &
         28.5 & \textbf{25.5} & \textbf{18.2} &  \textbf{5.3}&
        \textbf{27.3} & 9.9 & \textbf{10.0} & \textbf{1.8} \\
        \end{tabular}}
        \caption{MIT-States}
        \label{tab:mitstates}
        \end{subtable}
    \vfill
    \renewcommand\thetable{(b)} 
    \begin{subtable}[h]{1\linewidth}
    \centering
    \resizebox{1\linewidth}{!}
    {
    \begin{tabular}{l|cc|cccc|cccc}
    \toprule
    \multicolumn{1}{l}{\textbf{Method}} & \multicolumn{2}{c}{\textbf{Concepts}} & \multicolumn{4}{c}{\textbf{Close-World}} &
    \multicolumn{4}{c}{\textbf{Open-World}}\\
    \toprule
    & State & Object  
    &Seen & Unseen  & HM & AUC 
    &Seen & Unseen  & HM & AUC \\
    \toprule
    AoP & 38.9 & 69.6 &
    59.8 & 54.2&  40.8 & 25.9 &
    50.9 &  34.2     & 29.4   & 13.7 \\
    LE+ & 41.2 & 69.2 &
    53.0 & 61.9& 41.0 & 25.7 &
    60.4  & 36.5 & 30.5  & 16.3 \\
    TMN & 40.8 & 69.9 &
    58.7 & 60.0 & 45.0 & 29.3 &
    55.9  &18.1 & 21.7 &  8.4 \\
    SymNet & 40.5 & 71.2 &
    53.3 & 57.9 &  39.2 & 23.9 &
    53.3   &44.6 &34.5   &18.5 \\
    CompCos&  44.7 & 73.5 &
    59.8 & 62.5 & 43.1 & 28.7 &
    \textbf{ 59.3} & {46.8} & {36.9} & {21.3} \\
    CGE & 45.0 & 73.9 &
    56.8 & \textbf{63.6} & 41.2 & 26.4 &
    55.3 & 46.2 & 38.5 & 21.6 \\
        \bottomrule
    \textbf{\ours} & \textbf{55.0} & \textbf{77.2} &
    \textbf{65.0} & 62.4 & \textbf{49.8} & \textbf{34.6}& 
    58.6 &\textbf{ 48.4} & \textbf{41.7} & \textbf{22.2} \\
    \end{tabular}}
    \caption{UT-Zappos}
    \label{tab:utzap50k}
    \end{subtable}
    \vfill
    \renewcommand\thetable{(c)} 
    \begin{subtable}[h]{1\linewidth}
    \centering
    \resizebox{1\linewidth}{!}
    {
    \begin{tabular}{l|cc|cccc|cccc}
    \toprule
    \multicolumn{1}{l}{\textbf{Method}} & \multicolumn{2}{c}{\textbf{Concepts}} & \multicolumn{4}{c}{\textbf{Close-World}} &
    \multicolumn{4}{c}{\textbf{Open-World}}\\
    \toprule
    & State & Object  
    &Seen & Unseen  & HM & AUC 
    &Seen & Unseen  & HM & AUC \\
    \toprule
    AoP & 8.3 & 12.5 &
    11.8 & 3.9 & 2.9 & 0.3 &
    20.2 & 0.9 & 3.1 & 0.1 \\
    LE+ & 7.4 & 15.6 &
    16.1  & 5.0 & 5.3 & 0.6 &
    20.9 & 0.9 & 3.1 & 0.1 \\
    TMN & 9.7  &20.5 &
    21.6 & 6.3 & 7.7 & 1.1 &
    -&-&-&- \\
    SymNet & 14.5  & 20.2 &
    25.2 & 9.2 & 9.8 & 1.8 &
    24.6 & 2.1 & 5.1 & 0.4\\
    CompCos & 9.5  & 27.5 &
    27.0 & 10.5 & 11.7 & 2.3 &
    25.9 & 1.8 & 4.8 & 0.4 \\
    CGE & 12.7 &26.9 &
    27.5 & 11.7 & 11.9 & 2.5 &
    26.0 & 1.4 & 4.0 & 0.2 \\
    \bottomrule
    \textbf{\ours} & \textbf{28.2} & \textbf{34.9} &
    \textbf{28.2} & \textbf{11.9} & \textbf{13.9} & \textbf{2.8} &
    \textbf{26.6} & \textbf{2.9} & \textbf{6.4} & \textbf{0.7} \\
    \end{tabular}}
    \caption{\textbf{C-GQA}}
    \label{tab:cgqa}
    \end{subtable}
    \setcounter{table}{1}
    \renewcommand\thetable{2} 
    \renewcommand{\tablename}{Table}
    \caption{Comparison of performance on benchmark datasets (in \%): We report state and object accuracy of the primitive concepts; area under the curve (\auc), best seen, unseen and harmonic mean (HM) accuracies of the compositional pairs in both close-world and open-world settings. Best performance is in bold.}
    \vspace{-1.5em}
    \label{tab:CZSL_results}
\end{table}}

\subsection{Discussion of Results: CZSL Tasks}\label{results_CZSL}
We compare \ours with the state of the art methods on three benchmark datasets, i.e. MIT-States, UT-Zap50k and C-GQA, for both close world and open world settings. It can be seen from Table \ref{tab:CZSL_results} that \ours outperforms all other methods and establishes a new state-of-the-art for both close world and open world CZSL tasks.

\myparagraph{Closed World Results.} 
We observe from Table~\ref{tab:CZSL_results}\ref{tab:mitstates}) that \ours achieves a leading 5.3 \auc and 18.2 on the best harmonic mean metric on the MIT-States dataset.
We can also observe comparable performance on other metrics.
We note that even though CGE exhibit a higher accuracy on the state and seen composition predictions, \ours achieve an overall better results, thanks to more balanced predictions between unseen and seen compositions, as indicated by the harmonic mean metric. 
On the UT-Zappos dataset (see Table~\ref{tab:CZSL_results}\ref{tab:utzap50k}), we observe a huge performance improvement by \ours over SOTA methods.
\ie \auc and harmonic mean show an significant improvement of 18\% (29.3 to 34.6) and 10.7\% (45 to 49.8) respectively, over the second best method, TMN.
Similarly, we note 
from Table~\ref{tab:CZSL_results}\ref{tab:cgqa}.
that on C-GQA dataset significant performance gains are achieved by our approach in primitive concepts as well as compositional pairs.  
These consistent performance gains supports on all 3 datasets support our claim that
by adopting the variational approach \ours better captures the underlying data distribution than current SOTA methods.
\\

\myparagraph{Open World Results.}
The challenge of the open world setting can be seen clearly from the performance degradation experienced by all the models (see right side of Table~\ref{tab:CZSL_results}).
Due the the significantly huge search space, the accuracy on unseen compositions, and as a consequence also the \auc and the best harmonic mean, decreases rapidly. 
On both MIT-States and C-GQA,
the performance of the models is halved  on these three metrics. On the other hand, the impact is relatively milder for the UT-Zappos dataset. 
This is because the percentage of hypothetical pairs in the OW setting is just 40\% for UT-Zappos, whereas it is $\sim$93\% and $\sim$98\% for MIT-States and CGQA respectively (see Table~\ref{tab:Datasets}).

We observe from the results that
\ours also exhibits superior performance in the open world setting on all 3 datasets.
On MIT-States, with respect to best HM metric, \ours achieves 15\% and 12.5\%  improvement over CGE and Compcos respectively. 
Similar performance trend can be observed on UT-Zappos and C-GQA datasets, where \ours outperforms other models on \auc and best HM metric. The performance gains on the most challenging dataset for OW setting, \ie C-GQA, are the most promising.
For instance, \ours improves the best HM by 8.3\% and 25.5\% in comparison to the second best method on UT-Zappos and C-GQA datasets, respectively. 
This strengthens our claim that by adopting a variational graph based approach and focusing on learning good representation of primitive concepts in a disentangled way leads to better compositional generalization.

One interesting observation is that, on C-GQA dataset, \ours experiences a huge 
performance degradation of 75\% when 
comparing its \auc with CW setting. 
Whereas, CGE experiences a more dramatic performance degradation of 92\%.
This shows that even when the number of hypothetical pairs is ~98\% (\ie C-GQA, see Table~\ref{tab:Datasets}), \ours is quite effective relative to SOTA methods, in tackling the infeasible pairs in the open world setting.
This empirically shows the validity of our hypothesis that once good representation of primitive concepts are learnt in a disentangled way, they are more likely to be stable across seen, unseen and infeasible compositions.
Thus, such node embeddings and edges between the primitive concepts are sufficient for achieving good compositional generalization.
{
\begin{table}
    \centering
    \resizebox{0.85\linewidth}{!}{
    \begin{tabular}{l| c c   |  c c }
    \toprule
        \multicolumn{1}{l}{\textbf{Method}} & \multicolumn{2}{c}{\textbf{MIT-States}} & \multicolumn{2}{c}{\textbf{UT-Zappos}}
    \\\toprule
  &R@1 & R@10   & R@1 & R@10  \\\toprule
   {ComposeAE}
     & 13.9  & 47.9
     & 64.2 & 69.1 \\
     \toprule
    AoP            
     & 8.8 & 39.1 & 
     24.2 & 59.4 
     \\
    SymNet            
     & 11.2 & 41.4 &
     37.4 & 62.8 
     \\
     CompCos            
     & 17.5 & 47.3 &
     42.7 & 63.9 
     \\
     CGE           
     & 9.4 & 35.9 
     & 59.7 & 67.7 
     \\
     \bottomrule
    \textbf{\ours}            
     & \textbf{17.7} & \textbf{49.7} &
     \textbf{65.4} & \textbf{70.3} 
     \\

    \end{tabular}}
    \caption{Image retrieval results on MIT states and UT Zappos}
    \vspace{-18pt}
    \label{tab:ret}
\end{table}
}

\subsection{Discussion of Results for the Image Retrieval Based on Multi-modal Query Task} 
\label{results_RET}
In order to gain further evidence that the embeddings learnt via \ours capture the underlying data distribution better than current SOTA methods, we consider image retrieval task.
Recently, this task has been well-studied in the literature. 
The task is inspired from the potential application scenario, presented in Fig. 1. \ie, a user sends a multi-modal query (an image with a text request for some change in the attributes) and the task is to retrieve the images with desired modifications from the database. 
In the literature, this task is simplified by limiting the text request to only one word attribute (state) change.
ComposeAE \cite{anwaar2020compositional} is the current state-of-the-art method for this task.
We follow the same evaluation protocol as ComposeAE \cite{anwaar2020compositional}.
From \autoref{tab:ret}, we observe that \ours outperforms not only the competitive methods (\ie CompCos and CGE) but also ComposeAE on both datasets.
Specifically, on MIT-States and UT-Zappos, \ours achieves a Recall at position 1 (R@1) of 17.7 and 65.4, whereas CGE achieves 9.4 and 59.7 respectively. 
Similarly for R@10, the performance improvement over all the methods (including ComposeAE) is quite significant.

\begin{figure}[!h]
    \centering
    \includegraphics[width=0.99\linewidth]{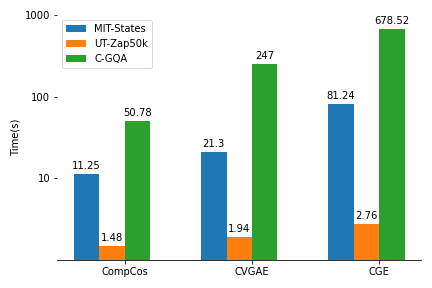}
    \caption{Comparison of training times (one epoch) of top-3 approaches on the OW-CZSL task. The y-axis is in log scale.
    We can see that \ours is quite time-efficient than its direct graph-based competitor, CGE. }
    \vspace{-1em}
    \label{fig:times}
\end{figure}
\begin{figure*}
\centering
\includegraphics[width=0.99\linewidth]{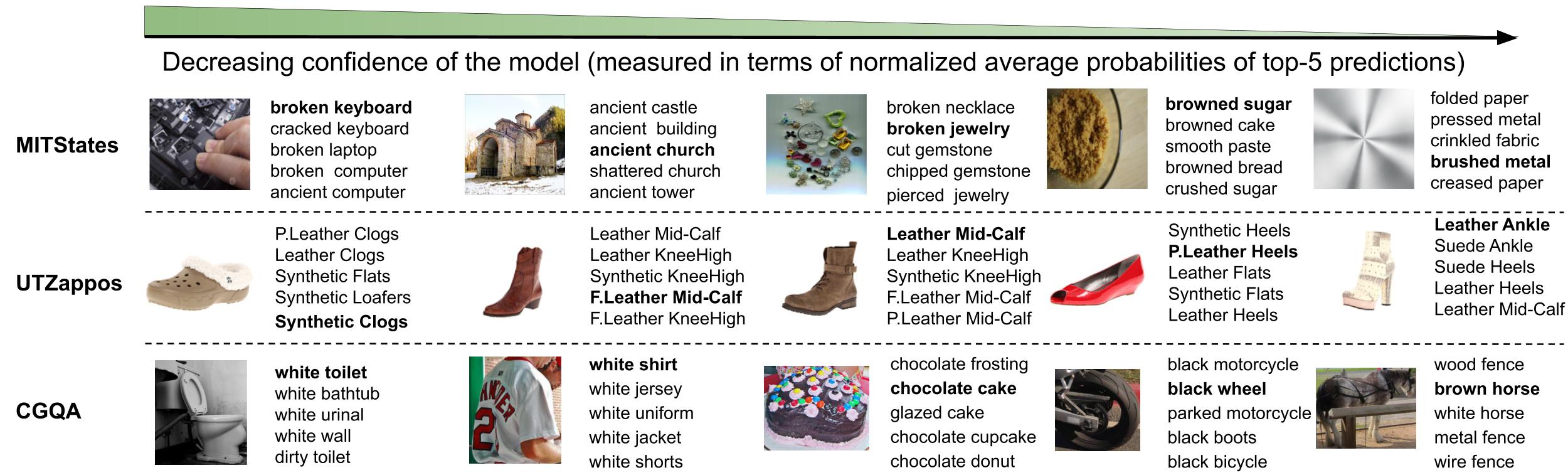}
\caption{Qualitative Retrieval Results: Top-5 Predictions of \ours. Bold indicates the true compositional label.}
\label{fig:Retrieval}
\end{figure*}

\subsection{Computational Complexity} \label{sec: time}
We now compare the computational complexity of \ours with the graph-based SOTA method CGE.
The graph in \ours consists of $N$ nodes, where $N=|\mathcal{S}| + |\mathcal{O}|$.
Whereas the graph proposed in CGE  \cite{cge}, consists of $N_{CGE}=|\mathcal{S}| + |\mathcal{O}| + |\mathcal{Y}_\xi|$ nodes. 
In general, 
$|\mathcal{Y}_\xi|\gg|\mathcal{S}| + |\mathcal{O}|$, for instance, in the OW setting, $|\mathcal{Y}_\xi| = |\mathcal{S}| \times |\mathcal{O}|$.
Such formulation makes CGE model computationally more expensive than \ours.
Formally, we assume that the graph is sparse with the number of edges $|\mathcal{E}|=\mathcal{O}(N)$ and each node is represented by $m$-dimensional features. Thus, the time and memory complexity for a $L$-layered GCN model can be written as $\mathcal{O}(L N m^2)$ and $\mathcal{O}(L N m + L m^2)$ respectively.
We can clearly see that \ours will be much more efficient than CGE, since $|N_{CGE}|\gg|N|$ for all real-world datasets.
\eg for C-GQA dataset, CGE requires \num{3.94e5} nodes and \num{1.07e4} nodes for OW and CW setting, respectively. While \ours requires only 1323 nodes for both settings.
It is to be noted that the C-GQA dataset contains only 453 states and 870 objects, whereas any real-world application scenario will involve much more number of states and objects.
Thus, CGE fails to tackle the problem of scalability.

We also empirically compare the training times of top-3 performing algorithms in \figgref{fig:times} for the OW-CZSL task.
As some of the methods (\eg CGE) are more resource intensive than others, we select AWS instance type \texttt{g4dn.4xlarge\footnote{\url{https://aws.amazon.com/ec2/instance-types/}}} for fair comparison of training times.
All hyperparameters are kept the same, \eg batchsize
is set to 128, for all three methods.
As expected, since Compcos does not employ any graph, so it is the fastest among the three methods.
We observe that between the two graph methods, \ours is quite time efficient. \eg for MIT-States, it is $\sim$4 times faster than CGE.
This provides empirical evidence for lower computational complexity of \ours as discussed above.

\subsection{Ablation Studies} \label{sec: ablation}
We have conducted various ablation studies, in order
to gain insight into the performance of \ours. 
Table \ref{tab:ablation_cgqa} presents the quantitative results of these studies.
\begin{table}
    \centering
    {
    \begin{tabular}{l|cc|cc|cc}
    \toprule
    \multicolumn{1}{l}{\textbf{Method}} & \multicolumn{2}{c}{\textbf{Concepts}} & \multicolumn{2}{c}{\textbf{Close-World}} &
    \multicolumn{2}{c}{\textbf{Open-World}}\\
    \toprule
    & State & Object  & HM & AUC & HM & AUC \\
    \toprule
    CGAE & {24.7} & {30.3} &
    {11.4} & {2.5} & {4.2} & {0.4} \\
    \bottomrule
    \textbf{\ours}$_{GAT}$ & {27.4} & {33.2} &
    {12.8} & {2.3} & {5.8} & {0.6} \\
    \textbf{\ours}$_{GCN}$ & {27.9} & {33.7} &
    {13.5} & {2.6} & {6.1} & {0.6} \\
    \textbf{\ours}$_{GCN-II}$ & {27.6} & {33.4} &
    {13.1} & {2.5} & {5.9} & {0.6} \\
    \bottomrule
    \textbf{\ours} & \textbf{28.2} & \textbf{34.9} &
    \textbf{13.9} & \textbf{2.8} & \textbf{6.4} & \textbf{0.7} \\
    \end{tabular}}
    \caption{Impact of Graph Encoder and Non-Variational approach on the performance of \ours on the C-GQA dataset}
    \label{tab:ablation_cgqa}
    \end{table}

\myparagraph{Non-variational approach:}
\ours follows a variational approach to learn the underlying data distributions of the concepts. 
Instead of VGAE, we used a non-variational GAE to quantify the effect of using variational approach on the performance. 
Row 1 in Table 4 shows that there is a drop in performance for both the CW and OW settings. 
This strengthens our hypothesis that it is better to
employ a variational approach for learning the representation of concepts. This especially aids in the OW setting, where the drop in performance is most significant. e.g., on AUC metric, there is a performance degradation of 42.8\%.

\begin{table}
    \centering
    {
    \begin{tabular}{l|c|c|c}
    \toprule
    \multicolumn{1}{l}{\textbf{$k$}} & \multicolumn{1}{c}{\textbf{MIT-States}} & \multicolumn{1}{c}{\textbf{UT-Zappos}} &
    \multicolumn{1}{c}{\textbf{C-GQA}}\\
    \toprule
    &  HM [CW/OW]  & HM [CW/OW] & HM [CW/OW] \\
    \toprule
    400 & {16.9/9.2} & {46.2/37.3} &
    {13.6/6.1} \\
    800 & {17.8/9.6} & {46.7/38.1} &
    \textbf{13.9/6.4} \\
    1200 & {17.9/9.8} & {48.6/38.9} &
    {13.5/6.2}  \\
    1600 & \textbf{18.2/10.0} & \textbf{49.8/41.7} &
    {13.2/5.7} \\
    2000 & {18.1/9.7} & {49.5/40.8} &
    {13.1/5.8} \\
    \end{tabular}}
    \caption{Effect of embedding dimension of common embedding space ($k$) on the performance of \ours measured in terms of best Harmonic Mean (HM) in the Close-World (CW) and Open-World (OW) settings}
    \label{tab:ablation_cgqa_dim}
    \end{table}

\myparagraph{Impact of graph encoder:}
We now look at the impact of the graph encoder on the performance of \ours.
We employ three graph encoders, namely: Graph Attention Network (GAT) \cite{velivckovic2017graph}, Graph Convolution Network (GCN) \cite{gcn} and its recent improved version (GCN-II) \cite{gcnii}.
Due to space constraints, we show the results on the C-GQA dataset only.
Table \ref{tab:ablation_cgqa} shows the effect of graph encoder on the performance of \ours.
We can observe that there is no significant decrease in the performance. The performance of GCN variants is consistently better than attention based graph encoder (GAT).

\myparagraph{Common Embedding Space:} 
Table \ref{tab:ablation_cgqa_dim} shows the effect of dimensions of the common embedding space on the performance of the model. In the interest of space, we only report best Harmonic Mean (HM) for the Close World (CW) and Open World (OW) settings. Rest of the metrics also follow a similar trend in performance.
We observe from the results that the choice of embedding dimension has a significant impact on the performance for all three datasets. Specifically, the performance degradation is 7.1\%, 7.2\% and 5.6\% with respect to CW-HM metric on MIT-States, UT-Zappos and C-GQA datasets respectively.

\subsection{Qualitative Results and Important Limitations} \label{sec: qual}
Fig.~\ref{fig:Retrieval} presents some
qualitative retrieval results for three datasets.
Each row represents image samples from one dataset and each column contains the top-5 composition label predictions by \ours for an image.
First we observe that the model is able to retrieve the compositional labels which share similar semantics. For instance, ``ancient church" is actually a subset of ``ancient building" in a conceptual world.
Similarly, ``broken keyboard" and ``cracked keyboard" capture same semantic meaning.
Second, we note that although model is able to predict several correct compositional labels but model is rewarded only if it predicts the label as annotated in the dataset. 
We argue that the task of CZSL is inherently multi-label task and benchmark datasets should be amended to cater for this aspect. \ie prediction should be considered true if the model predicts any of the compositional concepts present in the image. 
Otherwise, the model has to learn the annotator bias to perform well. For instance, the rightmost image in the third row has both ``wood fence" and ``brown horse" in it. Similarly, second image in the first row can be seen as both ``ancient building" and ``ancient church".

\myparagraph{Limitations:}
The journey towards achieving compositional generalization is an arduous one. 
Our work tries to solve part of the problem and get rid of some prior assumptions. 
But there is still a strong assumption that composition can be only done with two concepts.
This is quite limiting in practice. For instance, the approaches discussed in this work can not directly compose ``old red car". Although, in comparison to CGE \cite{cge}, \ours can be extended relatively easy to more than two concepts \ie by concatenating the graph embeddings of all three concepts of interest.
Furthermore, all the datasets considered also assume that there exists only two kinds of primitive concepts \ie objects and their states. There is a need of new datasets which consider more kinds of primitive concepts.
Another limitation of current work can be observed 
in Fig.~\ref{fig:Retrieval} \ie the confidence of model is not the same in predicting the composition of different concepts. The robustness of predictions across different concepts plays a crucial role in the real-world application of the model.
This aspect needs to be investigated further in light of recent works \cite{ross2018improving,abdar2021review,hendrycks2019augmix}.  
Another important limitation of our work is that we did not take into account the label noise in the datasets.
For instance, we can see that there are certain images, like the two rightmost images in the first row of Fig.~\ref{fig:Retrieval}, which are difficult for even humans to uniquely annotate.
Such issues hinder the model from achieving a good prediction accuracy. 

\section{Conclusion}
In this work, we proposed a variational graph autoencoder based approach to learn composition of objects and states,
in both close-world and open-world settings.
\ours learns the variational embeddings of the primitive concepts and treats the edge between them as proxy for the feasibility of the compositions.
The proposed approach outperforms the current state-of-the-art methods by significant margin on three widely-used benchmark datasets.
Based on our novel formulation of the problem, \ours
remains computationally cheaper than the SOTA method, CGE \cite{cge}.
The performance on the image retrieval task validates our claim that \ours learns better representations of primitive concepts than current SOTA methods.
We also present some qualitative results and discuss the limitations of current algorithms and datasets which hinders them from achieving compositional generalization.

\section*{Acknowledgments}

This work has been supported by the
Bavarian Ministry of Economic Affairs, Regional Development and Energy through the \emph{WoWNet} project  IUK-1902-003// IUK625/002.

{\small
\bibliographystyle{ieee_fullname}
\bibliography{main}

\begin{thebibliography}{10}\itemsep=-1pt

\bibitem{abdar2021review}
Moloud Abdar, Farhad Pourpanah, Sadiq Hussain, Dana Rezazadegan, Li Liu,
  Mohammad Ghavamzadeh, Paul Fieguth, Xiaochun Cao, Abbas Khosravi, U~Rajendra
  Acharya, et~al.
\newblock A review of uncertainty quantification in deep learning: Techniques,
  applications and challenges.
\newblock {\em Information Fusion}, 2021.

\bibitem{alhalah16}
Ziad Al-Halah, Makarand Tapaswi, and Rainer Stiefelhagen.
\newblock Recovering the missing link: Predicting class-attribute associations
  for unsupervised zero-shot learning.
\newblock In {\em 2016 IEEE Conference on Computer Vision and Pattern
  Recognition (CVPR)}, pages 5975--5984, 2016.

\bibitem{anwaar2020compositional}
Muhammad~Umer Anwaar, Egor Labintcev, and Martin Kleinsteuber.
\newblock Compositional learning of image-text query for image retrieval.
\newblock In {\em Proceedings of the IEEE/CVF Winter Conference on Applications
  of Computer Vision (WACV)}, pages 1140--1149, January 2021.

\bibitem{PsychologicalBiederman}
Irving Biederman.
\newblock Recognition-by-components: a theory of human image understanding.
\newblock {\em Psychological review}, 94 2:115--147, 1987.

\bibitem{pos1}
Bernd Bohnet, Ryan McDonald, Gon{\c{c}}alo Sim{\~o}es, Daniel Andor, Emily
  Pitler, and Joshua Maynez.
\newblock Morphosyntactic tagging with a meta-{B}i{LSTM} model over context
  sensitive token encodings.
\newblock In {\em Proceedings of the 56th Annual Meeting of the Association for
  Computational Linguistics (Volume 1: Long Papers)}, pages 2642--2652,
  Melbourne, Australia, July 2018. Association for Computational Linguistics.

\bibitem{fasttext}
Piotr Bojanowski, Edouard Grave, Armand Joulin, and Tomas Mikolov.
\newblock Enriching word vectors with subword information.
\newblock {\em Transactions of the Association for Computational Linguistics},
  5:135--146, 2017.

\bibitem{chao2017empirical}
Wei-Lun Chao, Soravit Changpinyo, Boqing Gong, and Fei Sha.
\newblock An empirical study and analysis of generalized zero-shot learning for
  object recognition in the wild, 2017.

\bibitem{chen2014inferring}
Chao-Yeh Chen and Kristen Grauman.
\newblock Inferring analogous attributes.
\newblock In {\em CVPR}, 2014.

\bibitem{choi2013adding}
Jonghyun Choi, Mohammad Rastegari, Ali Farhadi, and Larry~S Davis.
\newblock Adding unlabeled samples to categories by learned attributes.
\newblock In {\em CVPR}, 2013.

\bibitem{deng2014large}
Jia Deng, Nan Ding, Yangqing Jia, Andrea Frome, Kevin Murphy, Samy Bengio, Yuan
  Li, Hartmut Neven, and Hartwig Adam.
\newblock Large-scale object classification using label relation graphs.
\newblock In {\em ECCV}, 2014.

\bibitem{deng2009imagenet}
Jia Deng, Wei Dong, Richard Socher, Li-Jia Li, Kai Li, and Li Fei-Fei.
\newblock Imagenet: A large-scale hierarchical image database.
\newblock In {\em CVPR}, 2009.

\bibitem{vae}
Carl Doersch.
\newblock Tutorial on variational autoencoders.
\newblock {\em arXiv preprint arXiv:1606.05908}, 2016.

\bibitem{farhadi09}
Ali Farhadi, Ian Endres, Derek Hoiem, and David Forsyth.
\newblock Describing objects by their attributes.
\newblock In {\em 2009 IEEE Conference on Computer Vision and Pattern
  Recognition}, pages 1778--1785, 2009.

\bibitem{GraphSAGE}
William~L. Hamilton, Rex Ying, and Jure Leskovec.
\newblock Inductive representation learning on large graphs, 2018.

\bibitem{hendrycks2019augmix}
Dan Hendrycks, Norman Mu, Ekin~D Cubuk, Barret Zoph, Justin Gilmer, and Balaji
  Lakshminarayanan.
\newblock Augmix: A simple data processing method to improve robustness and
  uncertainty.
\newblock {\em arXiv preprint arXiv:1912.02781}, 2019.

\bibitem{CognitionHoffman1984PartsOR}
Donald~D. Hoffman and Whitman Richards.
\newblock Parts of recognition.
\newblock {\em Cognition}, 18:65--96, 1984.

\bibitem{isola2015mitstates}
Phillip Isola, Joseph~J Lim, and Edward~H Adelson.
\newblock Discovering states and transformations in image collections.
\newblock In {\em CVPR}, 2015.

\bibitem{jayaraman14ua}
Dinesh Jayaraman and Kristen Grauman.
\newblock Zero-shot recognition with unreliable attributes.
\newblock In {\em Proceedings of the 27th International Conference on Neural
  Information Processing Systems - Volume 2}, NIPS'14, page 3464–3472, 2014.

\bibitem{jayaraman14}
Dinesh Jayaraman, Fei Sha, and Kristen Grauman.
\newblock Decorrelating semantic visual attributes by resisting the urge to
  share.
\newblock In {\em 2014 IEEE Conference on Computer Vision and Pattern
  Recognition}, pages 1629--1636, 2014.

\bibitem{kingma2014adam}
Diederik~P Kingma and Jimmy Ba.
\newblock Adam: A method for stochastic optimization.
\newblock {\em ICLR}, 2015.

\bibitem{vgae}
Thomas~N Kipf and Max Welling.
\newblock Variational graph auto-encoders.
\newblock {\em arXiv preprint arXiv:1611.07308}, 2016.

\bibitem{gcn}
Thomas~N Kipf and Max Welling.
\newblock Semi-supervised classification with graph convolutional networks.
\newblock In {\em ICLR}, 2017.

\bibitem{koushik2017}
Jayanth Koushik, Hiroaki Hayashi, and Devendra~Singh Sachan.
\newblock Compositional reasoning for visual question answering.
\newblock In {\em Proceedings of the 34 th International Conference on Machine
  Learning, 2017}, 2017.

\bibitem{kumar08}
Neeraj Kumar, Peter Belhumeur, and Shree Nayar.
\newblock Facetracer: A search engine for large collections of images with
  faces.
\newblock In {\em Computer Vision -- ECCV 2008}, pages 340--353, Berlin,
  Heidelberg, 2008. Springer Berlin Heidelberg.

\bibitem{lecun1989backpropagation}
Yann LeCun, Bernhard Boser, John~S Denker, Donnie Henderson, Richard~E Howard,
  Wayne Hubbard, and Lawrence~D Jackel.
\newblock Backpropagation applied to handwritten zip code recognition.
\newblock {\em Neural computation}, 1(4):541--551, 1989.

\bibitem{li2020symnet}
Yong-Lu Li, Yue Xu, Xiaohan Mao, and Cewu Lu.
\newblock Symmetry and group in attribute-object compositions.
\newblock In {\em CVPR}, 2020.

\bibitem{liu15}
Ziwei Liu, Ping Luo, Xiaogang Wang, and Xiaoou Tang.
\newblock Deep learning face attributes in the wild.
\newblock In {\em 2015 IEEE International Conference on Computer Vision
  (ICCV)}, pages 3730--3738, 2015.

\bibitem{compcos}
M Mancini, MF Naeem, Y Xian, and Zeynep Akata.
\newblock Open world compositional zero-shot learning.
\newblock In {\em 34th IEEE Conference on Computer Vision and Pattern
  Recognition}. IEEE, 2021.

\bibitem{word2vec}
Tomas Mikolov, Ilya Sutskever, Kai Chen, Greg~S Corrado, and Jeff Dean.
\newblock Distributed representations of words and phrases and their
  compositionality.
\newblock In {\em NIPS}, 2013.

\bibitem{gcnii}
Zhewei~Wei Ming~Chen, Bolin~Ding Zengfeng~Huang, and Yaliang Li.
\newblock Simple and deep graph convolutional networks.
\newblock In {\em ICML}, 2020.

\bibitem{misra2017redwine}
Ishan Misra, Abhinav Gupta, and Martial Hebert.
\newblock From red wine to red tomato: Composition with context.
\newblock In {\em CVPR}, 2017.

\bibitem{musgrave2020metric}
Kevin Musgrave, Serge Belongie, and Ser-Nam Lim.
\newblock A metric learning reality check, 2020.

\bibitem{cge}
MF Naeem, Y Xian, F Tombari, and Zeynep Akata.
\newblock Learning graph embeddings for compositional zero-shot learning.
\newblock In {\em 34th IEEE Conference on Computer Vision and Pattern
  Recognition}. IEEE, 2021.

\bibitem{nagarajan2018attributeasoperators}
Tushar Nagarajan and Kristen Grauman.
\newblock Attributes as operators: factorizing unseen attribute-object
  compositions.
\newblock In {\em ECCV}, 2018.

\bibitem{parikh11}
D. Parikh and K. Grauman.
\newblock Relative attributes.
\newblock In {\em 2011 International Conference on Computer Vision (ICCV)},
  2011.

\bibitem{patricia2014learning}
Novi Patricia and Barbara Caputo.
\newblock Learning to learn, from transfer learning to domain adaptation: A
  unifying perspective.
\newblock In {\em CVPR}, 2014.

\bibitem{pennington2014glove}
Jeffrey Pennington, Richard Socher, and Christopher~D Manning.
\newblock Glove: Global vectors for word representation.
\newblock In {\em Proceedings of the 2014 conference on empirical methods in
  natural language processing (EMNLP)}, pages 1532--1543, 2014.

\bibitem{purushwalkam2019tmn}
Senthil Purushwalkam, Maximilian Nickel, Abhinav Gupta, and Marc'Aurelio
  Ranzato.
\newblock Task-driven modular networks for zero-shot compositional learning.
\newblock In {\em ICCV}, 2019.

\bibitem{ross2018improving}
Andrew~Slavin Ross and Finale Doshi-Velez.
\newblock Improving the adversarial robustness and interpretability of deep
  neural networks by regularizing their input gradients.
\newblock In {\em Thirty-second AAAI conference on artificial intelligence},
  2018.

\bibitem{salakhutdinov2011learning}
Ruslan Salakhutdinov, Antonio Torralba, and Josh Tenenbaum.
\newblock Learning to share visual appearance for multiclass object detection.
\newblock In {\em CVPR}, 2011.

\bibitem{pos2}
Cicero~Dos Santos and Bianca Zadrozny.
\newblock Learning character-level representations for part-of-speech tagging.
\newblock In Eric~P. Xing and Tony Jebara, editors, {\em Proceedings of the
  31st International Conference on Machine Learning}, volume~32 of {\em
  Proceedings of Machine Learning Research}, pages 1818--1826, Bejing, China,
  22--24 Jun 2014. PMLR.

\bibitem{velivckovic2017graph}
Petar Veli{\v{c}}kovi{\'c}, Guillem Cucurull, Arantxa Casanova, Adriana Romero,
  Pietro Lio, and Yoshua Bengio.
\newblock Graph attention networks.
\newblock {\em arXiv preprint arXiv:1710.10903}, 2017.

\bibitem{TIRG}
Nam Vo, Lu Jiang, Chen Sun, Kevin Murphy, Li-Jia Li, Li Fei-Fei, and James
  Hays.
\newblock Composing text and image for image retrieval - an empirical odyssey.
\newblock In {\em Proceedings of the IEEE/CVF Conference on Computer Vision and
  Pattern Recognition (CVPR 2019)}, June 2019.

\bibitem{wang2017learning}
Yu-Xiong Wang, Deva Ramanan, and Martial Hebert.
\newblock Learning to model the tail.
\newblock In {\em NIPS}, 2017.

\bibitem{jensen}
Eric~W. Weisstein.
\newblock Jensen's inequality. {From MathWorld---A Wolfram Web Resource}.

\bibitem{zslxian18benchmark}
Yongqin Xian, Christoph~H Lampert, Bernt Schiele, and Zeynep Akata.
\newblock Zero-shot learning—a comprehensive evaluation of the good, the bad
  and the ugly.
\newblock {\em IEEE TPAMI}, 41(9):2251--2265, 2018.

\bibitem{yu2017zappos}
Aron Yu and Kristen Grauman.
\newblock Semantic jitter: Dense supervision for visual comparisons via
  synthetic images.
\newblock In {\em CVPR}, 2017.

\bibitem{zeiler2014visualizing}
Matthew~D Zeiler and Rob Fergus.
\newblock Visualizing and understanding convolutional networks.
\newblock In {\em ECCV}, 2014.

\end{thebibliography}
}

\clearpage
\appendix
\section{Supplementary Material: Derivation of ELBO Bound} \label{ELBO Bound}
We aim to learn the free parameters of our model such that the log probability of $\mathcal{G}$ is maximized \ie
{\small
\begin{align}
	log\Big(p(\mathcal{G})\Big) & = log\Big(\int{ p( \bm{Z}) p_{\theta}(\mathcal{G}| \bm{Z}) \ d\bm{Z}}\Big) \nonumber                                                                                                    \\
	                            & = log\Big(\int{ \frac{q_{\phi}(\bm{Z} | \mathcal{G})}{q_{\phi}(\bm{Z} | \mathcal{G})} p( \bm{Z}) p_{\theta}(\mathcal{G}| \bm{Z}) \ d\bm{Z}}\Big) \nonumber                                            \\
	                            & = log\Big(\mathbb{E}_{\bm{Z} \sim q_{\phi}(\bm{Z} | \mathcal{G})} \Big\{\frac{p( \bm{Z}) p_{\theta}(\mathcal{G}| \bm{Z})}{q_{\phi}(\bm{Z} | \mathcal{G})}\Big\}\Big) , \label{eq:vgar-obj-before-jensen}
\end{align}
}%

In order to ensure computational tractability, we use Jensen's Inequality \cite{jensen} to get ELBO bound of Eq.~\eqref{eq:vgar-obj-before-jensen}. i.e.

{\small
\begin{align}
	log\Big(p(\mathcal{G})\Big) & \geq \mathbb{E}_{\bm{Z} \sim q_{\phi}(\bm{Z} | \mathcal{G})} \Big\{ log\Big(\frac{p( \bm{Z}) p_{\theta}(\mathcal{G}| \bm{Z})}{q_{\phi}(\bm{Z} | \mathcal{G})}\Big)\Big\}                                                                                            \\
	                            & = \mathbb{E}_{\bm{Z} \sim q_{\phi}(\bm{Z} | \mathcal{G})} \Big\{ log\Big( p_{\theta}(\mathcal{G}| \bm{Z})\Big)\Big\} \nonumber \\ 
								& + \mathbb{E}_{\bm{Z} \sim q_{\phi}(\bm{Z} | \mathcal{G})} \Big\{ log\Big(\frac{p( \bm{Z}) }{q_{\phi}(\bm{Z} | \mathcal{G})}\Big)\Big\} \label{o-vgae}
\end{align}
}%

We follow Kipf \etal \cite{vgae} and restrict the decoder $p_{\theta}(\mathcal{G}| \bm{Z})$ to reconstruct only edge information from the latent space.
The edge information is contained in the adjacency matrix $\bm{A}$. In other words, we choose the decoder to be an edge decoder \ie $p_{\theta}(\bm{A}| \bm{Z})$. 
{\small
\begin{align}
log\Big(p(\mathcal{G})\Big) & \geq \mathbb{E}_{\bm{Z} \sim q_{\phi}(\bm{Z} | \mathcal{G})} \Big\{ log\Big( p_{\theta}(\bm{A}| \bm{Z})\Big)\Big\} \nonumber \\
	  & - D_{KL}\Big(q_{\phi}(\bm{Z} | \mathcal{G}) || p(\bm{Z}) \Big) \label{eq:vgae-obj-abstract}
	  \end{align}
}
where, $D_{KL}$ denotes the Kullback-Leibler (KL) divergence between the prior and approximate posterior distributions.  
By using \eqref{eq:vgae-pz}, \eqref{eq:vgae-pz_i}, \eqref{eq:vgae-qz_given_G} and \eqref{eq:vgae-qz_i_given_G}, the loss function can be formulated as negative of ELBO bound \eqref{eq:vgae-obj-abstract} \ie

{\small
\begin{align}
	\mathcal{L}_{\mathrm{ELBO}}  & = \sum \limits_{i=1}^{N}D_{KL}\bigg(\mathcal{N}\Big(\bm{\mu}_i(\mathcal{G}), \bm{\sigma}^2_i(\mathcal{G})\Big) \ || \  \mathcal{N}(\bm{0}, \mathrm{diag}(\bm{1})) \bigg) \nonumber \\
	  &  
	  - \mathbb{E}_{\bm{Z} \sim q_{\phi}(\bm{Z} | \mathcal{G})} \Big\{ log\Big( p_{\theta}(\bm{A}| \bm{Z})\Big)\Big\}. \nonumber
	  \label{eq:vgae-final-loss} 
\end{align}
}%
\clearpage

\end{document}